\documentclass[conference,a4paper]{IEEEtran}

\IEEEoverridecommandlockouts

\usepackage{cite}
\usepackage{amsmath,amssymb}
\usepackage{booktabs}
\usepackage{pifont}
\usepackage{xcolor}
\usepackage{microtype}
\usepackage[hidelinks]{hyperref}
\usepackage{orcidlink}
\usepackage{graphicx}
\usepackage{subcaption}
\graphicspath{{figures/}}

\renewcommand{\arraystretch}{1.05}

\usepackage[export]{adjustbox}

\newcommand{\Ours}{Ours}
\newcommand{\cmark}{\ding{51}}
\newcommand{\xmark}{\ding{55}}

\newcommand{\phbox}[2]{%
  \fbox{\parbox[c][0.40\linewidth][c]{#1}{\centering\scriptsize\ttfamily
  \color{gray}\detokenize{#2}}}}
\newcommand{\smartfig}[2][\linewidth]{%
  \IfFileExists{figures/#2}{\includegraphics[width=#1]{#2}}{\phbox{#1}{#2}}}
\begin{document}

\title{Multi-View Reflective Surface Inspection via Semantic--Saliency Cross-Verification}

\author{
\IEEEauthorblockN{
\mbox{
Van-Giang Nguyen\textsuperscript{\(\dagger\)},
Thanh-Tuan Tran\textsuperscript{\(\dagger\)}
\,\orcidlink{0009-0006-4370-8326},
Xuan-Hieu Phan
\,\orcidlink{0000-0002-7640-9190},
\text{and}
Xiem HoangVan\textsuperscript{*}
\,\orcidlink{0000-0002-7524-6529}
}
}

\IEEEauthorblockA{
\footnotesize
University of Engineering and Technology,
Vietnam National University, Hanoi 10000, Vietnam.\\[-0.1em]
\textsuperscript{\(\dagger\)}\textit{Equal contribution}
\qquad
\textsuperscript{*}\textit{Corresponding author}
}
}

\maketitle

\begin{abstract}
Reflective smartphone cover glass is challenging to inspect from a single fixed viewpoint because defect visibility varies with viewing geometry and specular reflections. This gives rise to two practical challenges: defects may be weakly observable from certain viewpoints, while the available visual evidence may remain spatially ambiguous. To address these issues, we propose a multi-view inspection framework in which each RGB observation is processed by a shared per-view expert. A vision--language model (VLM) produces
class-aware semantic boxes, while a normal-reference reconstruction branch
provides class-agnostic saliency. Their spatial agreement is used as supporting
evidence to rank semantic proposals without modifying their coordinates or
treating saliency as ground truth. The resulting evidence records are combined
at product level without cross-view registration. On 282 production-line
images, semantic--saliency association improves AP$_{50}$ from 52.6\% to
62.6\% by re-ranking fixed semantic proposals. Across 94 products, cross-view
evidence recall $R_{\rm prod}@0.5$ increases from 75.5\% for the best single
view to 88.3\% using all three views. These results support the complementary
roles of semantic--saliency cross-verification and additional optical
observations in reflective-surface inspection.
\end{abstract}


\section{Introduction}

Smartphone cover glass is difficult to inspect from a single fixed observation because defect visibility depends strongly on the imaging condition. The surface is dark and highly reflective, while defects such as scratches and cracks may occupy only a small region of the image. Their contrast changes with the geometry of illumination, surface, and sensor. Existing smartphone glass inspection systems therefore acquire the surface under different illumination directions or viewpoints to expose weak defects that may be inconspicuous in one observation \cite{turko2021smartphone,miao2025dual}. This motivates the use of multiple observations as complementary optical evidence for the same product.

Conventional inspection methods commonly formulate defect localization as supervised object detection. Faster R-CNN and recent YOLO variants provide strong localization when the target categories and representative box annotations are available \cite{ren2015faster,tian2025yolov12}. Their semantic scope, however, is determined by the defect classes represented during training. Changes in the inspection vocabulary therefore require corresponding labeled examples and model adaptation. This can be restrictive in manufacturing, where inspection criteria may include both local surface defects and relational conditions such as misalignment or unexpected components. Open-set grounding and vision-language models provide a more flexible semantic interface because the target concept can be specified through language \cite{liu2024groundingdino,gu2024anomalygpt,fan2025mavila}. This flexibility does not guarantee precise localization. Recent industrial vision-language studies still report difficulty in resolving small local anomalies and producing accurate anomaly localization \cite{gu2024anomalygpt,cheng2025anomalycot}.
\begin{figure}[!t]
\centering
\includegraphics[width=0.5\textwidth]{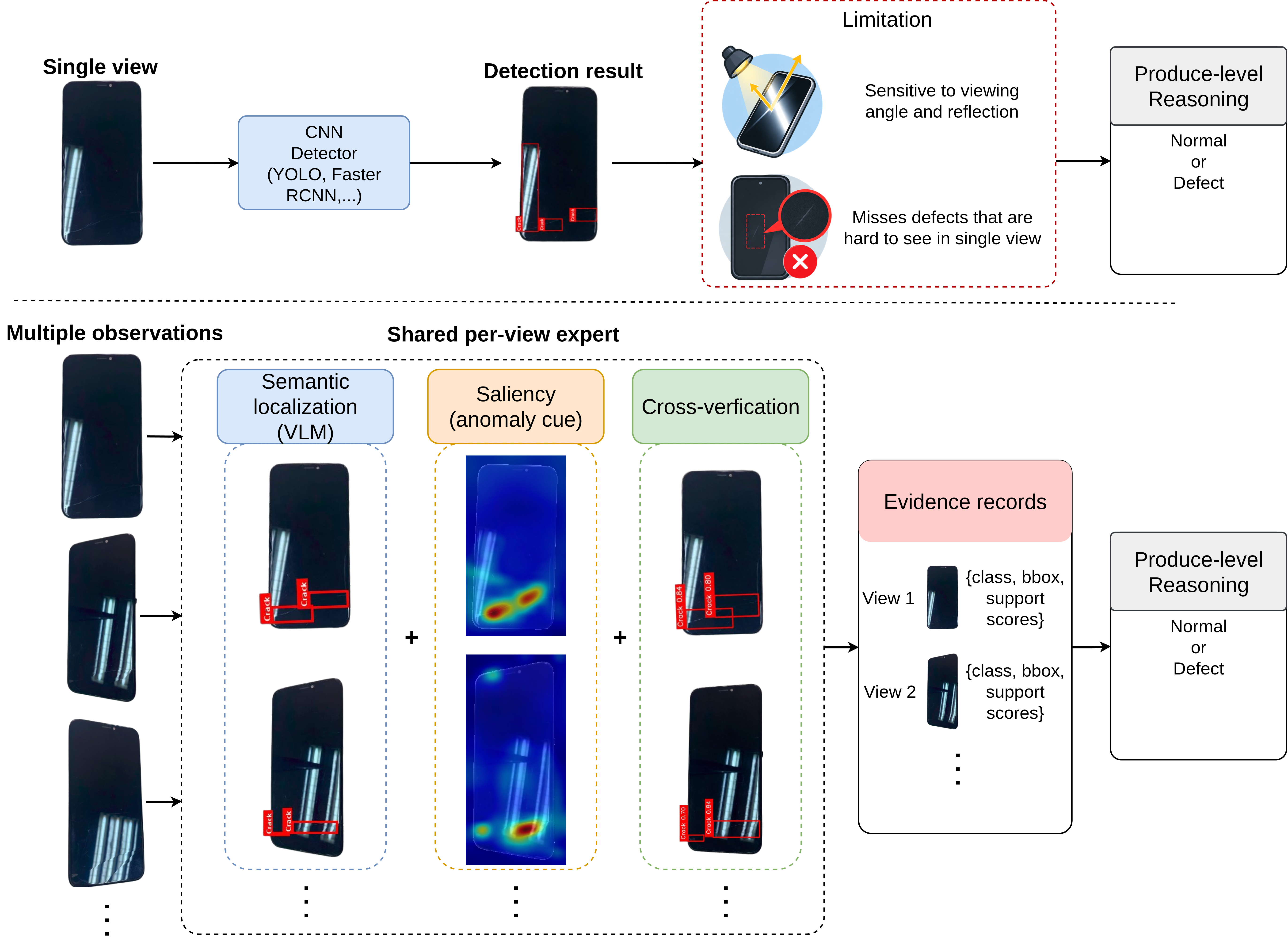}
\caption{Conventional single-view inspection (above) and the proposed multi-view semantic--saliency framework (below). Per-view semantic proposals are cross-verified with saliency evidence before product-level aggregation.}
\label{fig:usecase}
\end{figure}
Normal-reference anomaly detection provides a complementary source of spatial information. Embedding and reconstruction methods model normal appearance and localize regions that deviate from it without requiring labels for every defect category \cite{roth2022patchcore,zavrtanik2021draem,hicsonmez2026vlmdiff}. These responses are spatially informative but class-agnostic. On reflective glass, however, a strong appearance deviation may arise from either a true defect or a specular pattern. Saliency therefore cannot be treated as ground truth for a semantic prediction. Conversely, a semantic prediction may identify the correct defect concept while providing imprecise localization. We use the spatial agreement between semantic localization and separately generated normal-reference saliency as supporting evidence rather than allowing either branch to determine the result alone.

Multiple observations address a different limitation. Real-IAD established multi-view industrial anomaly evaluation, while subsequent work such as Multi-Flow explored information exchange across observations \cite{wang2024realiad,kruse2025multiflow}. Our objective is different. The views are used to expose the reflective surface under different optical conditions rather than to recover geometry or establish spatial correspondence across cameras. Each image is processed in its native coordinate system, and only the resulting evidence is combined at product level.

Fig.~\ref{fig:usecase} contrasts conventional single-view inspection with our formulation. Each RGB observation is processed by a shared per-view expert, where a VLM provides class-aware localization and a normal-reference branch provides spatial deviation evidence. Their association forms a per-view evidence record, and the resulting records are jointly considered at product level to determine the inspection outcome. The predicted defect and inspection context are then used to retrieve the corresponding manufacturer rule for reporting.

The main contributions are as follows.
\begin{itemize}
    \item We propose a multi-view inspection framework that aggregates per-view evidence at product level without cross-view registration.

    \item We introduce semantic-saliency cross-verification, where normal-reference saliency provides spatial support for VLM defect proposals without modifying their localization.

    \item We evaluate the per-view expert on a public benchmark and a conveyor-based production prototype, separately quantifying within-view association and cross-view optical complementarity.
\end{itemize}

\begin{figure*}[t]
\centering
\includegraphics[width=0.78\textwidth]{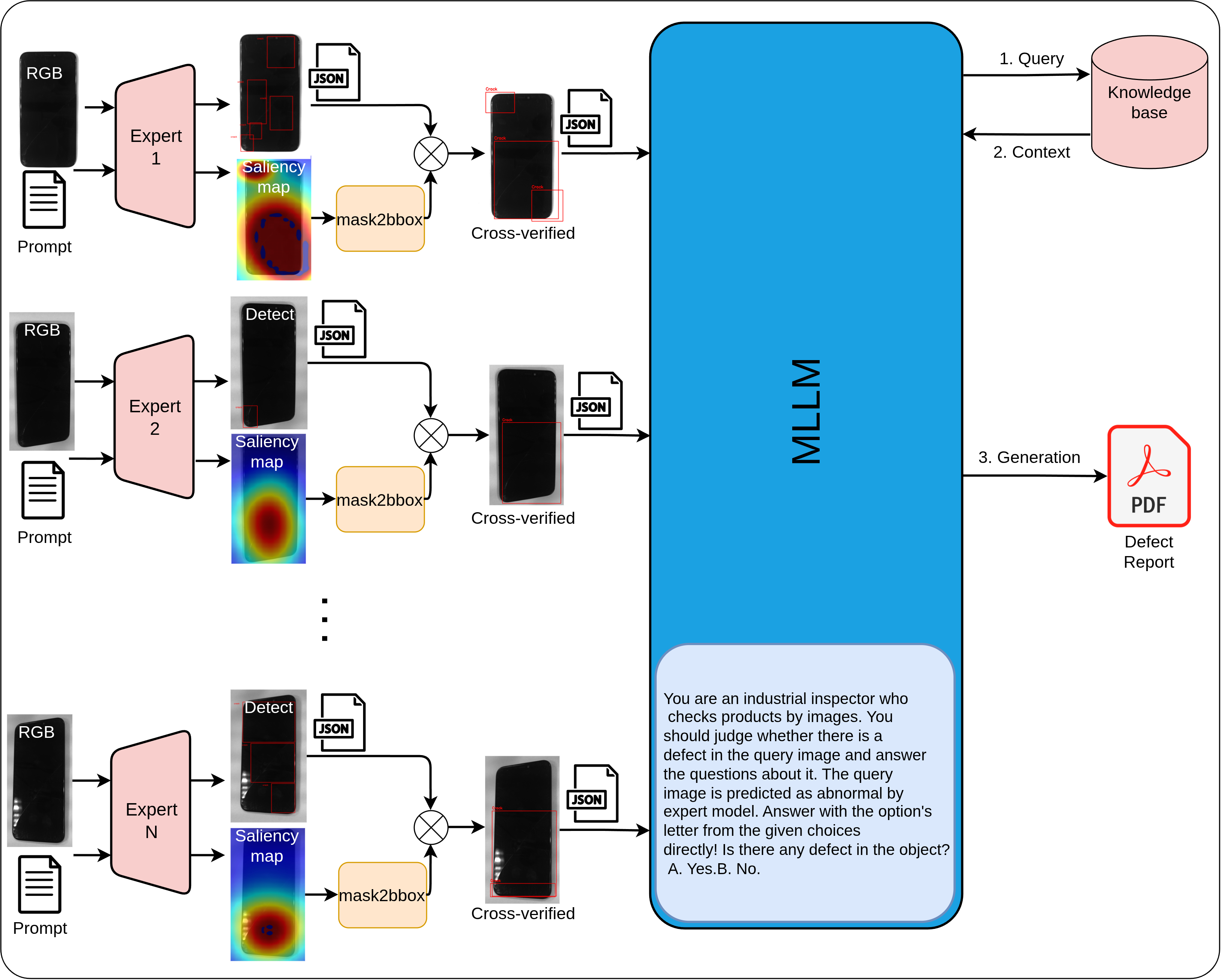}
\caption{System overview. Each RGB observation is processed by the same per-view expert, which associates VLM semantic proposals with normal-reference saliency as spatial support. The resulting evidence records are combined at product level without cross-view registration. Retrieval is applied after the visual verdict to populate the inspection report.}
\label{fig:pipeline}
\end{figure*}

\section{Method}

The proposed system, shown in Fig.~\ref{fig:pipeline}, separates per-view
evidence construction from product-level reasoning. Each observation is first converted into a common evidence record through semantic--saliency association, and only these records are combined
across views.

\subsection{Per-View Semantic--Saliency Expert}

\begin{figure}[t]
\centering
\includegraphics[width=0.98\linewidth]{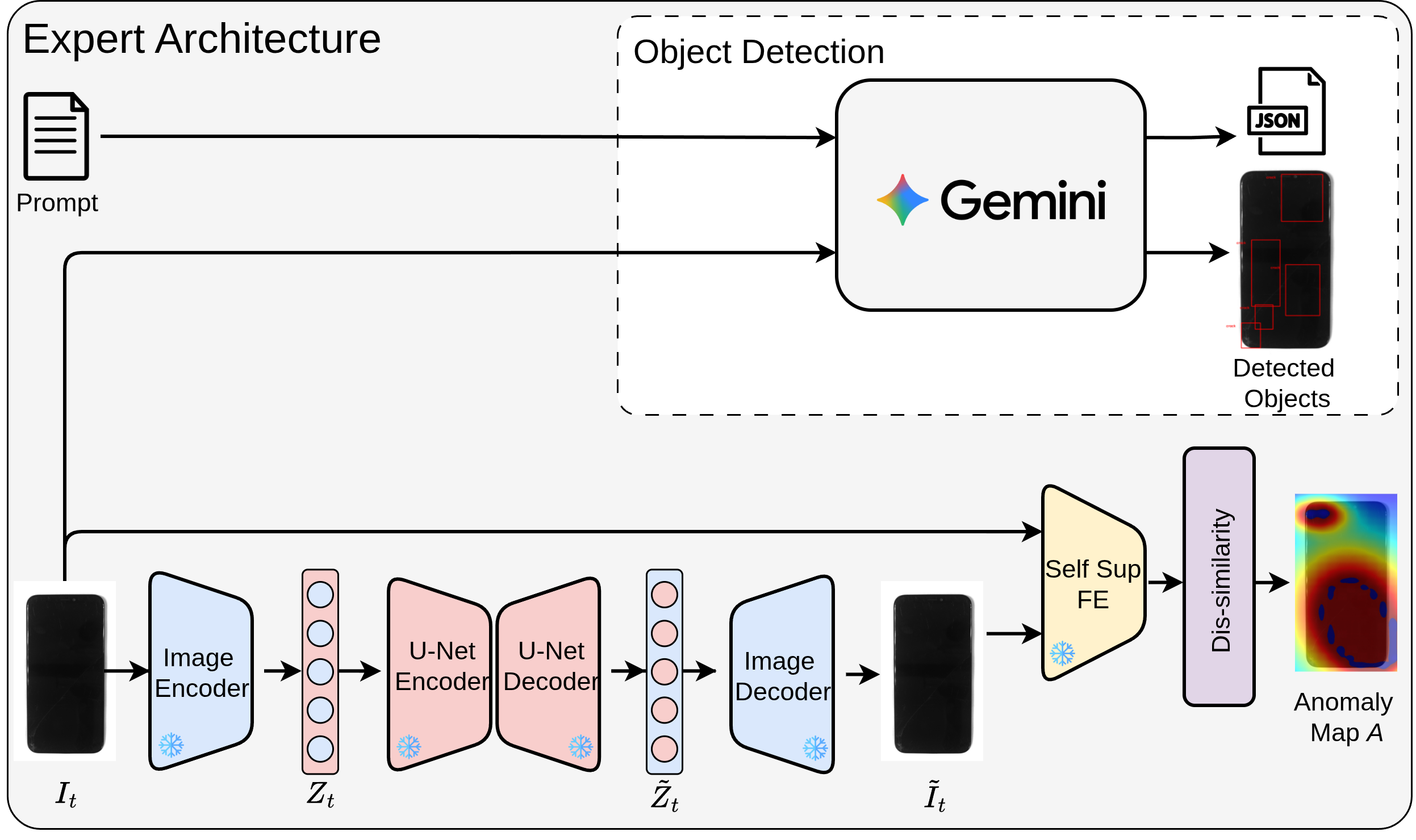}
\caption{Per-view semantic--saliency expert. The VLM produces class-aware semantic proposals, while the normal-reference branch provides class-agnostic saliency. Their spatial agreement is used as support while preserving the original semantic boxes.}
\label{fig:expert}
\end{figure}

Given an RGB observation $I_v$, the expert in Fig.~\ref{fig:expert} extracts two complementary cues. The VLM provides defect semantics and localization, while the normal-reference branch provides spatial evidence of deviation from expected appearance. The two prediction paths remain separate until spatial association.

\textbf{Semantic localization.}
The VLM receives $I_v$ and a fixed inspection prompt $P_{\rm det}$ specifying the defect vocabulary, relevant geometric cues, and output schema. Its parsed output is
\begin{equation}
\mathcal D_v=\{(c_{vj},b^s_{vj})\}_{j=1}^{n_v},
\label{eq:semantic}
\end{equation}
where $c_{vj}$ is the predicted class and $b^s_{vj}$ is the corresponding semantic box. Coordinates are generated on a normalized $[0,1000]$ grid and mapped deterministically to image pixels. Invalid classes and malformed boxes are rejected without an additional language-model call.

\textbf{Normal-reference evidence.}
A complementary spatial cue is obtained from the reconstruction pathway of VLMDiff \cite{hicsonmez2026vlmdiff}, adapted using normal target-domain images and frozen during evaluation. Let $\widehat I_v=g_\theta(I_v)$ denote the reconstructed image and $\Delta_v^{(\ell)}(x)$ the feature-space dissimilarity between $I_v$ and $\widehat I_v$ at level $\ell$. The responses are combined into
\begin{equation}
A_v(x)=\operatorname{Norm}\!\left(
\sum_{\ell=1}^{L} w_\ell
\operatorname{Up}\!\left(\Delta_v^{(\ell)}\right)(x)
\right),
\qquad
\sum_{\ell=1}^{L}w_\ell=1,
\label{eq:saliency}
\end{equation}
where $A_v$ is the normal-deviation saliency map. Thresholding $A_v$ at $\tau_a$ and removing components smaller than $\tau_{\rm area}$ yields
\begin{equation}
\mathcal A_v=\{b^a_{vk}\}_{k=1}^{m_v},
\label{eq:saliency_boxes}
\end{equation}
where each $b^a_{vk}$ bounds one surviving saliency region.

\textbf{Semantic--saliency cross-verification.}
The two branches produce different spatial representations. The VLM returns class-aware boxes, while the normal-reference branch produces a dense class-agnostic response. Connected saliency regions are converted to boxes only to place both cues in a common spatial representation, allowing their consistency to be measured directly. For semantic proposal $j$,
\begin{equation}
O_{vjk}=\operatorname{IoU}(b^s_{vj},b^a_{vk}),
\qquad
o^*_{vj}=\max_k O_{vjk},
\label{eq:association}
\end{equation}
with $o^*_{vj}=0$ when no saliency region is present. A proposal is marked as spatially supported when
\begin{equation}
u_{vj}=\mathbb I[o^*_{vj}\ge\tau_o].
\label{eq:support}
\end{equation}

The overlap $o^*_{vj}$ is used only as spatial support for ranking. It is not a calibrated confidence and does not establish prediction correctness. Semantic coordinates remain unchanged, and weakly supported proposals are retained because reflection can produce strong saliency while some relational defects may yield weak local residuals.

The per-view evidence is summarized as
\begin{equation}
\mathcal E_v=
\left(
\mathcal D_v,
\mathcal A_v,
\mathbf{o}_v,
\mathbf{u}_v,
F_v
\right),
\label{eq:record}
\end{equation}
where $\mathbf{o}_v=\{o^*_{vj}\}_{j=1}^{n_v}$, $\mathbf{u}_v=\{u_{vj}\}_{j=1}^{n_v}$, and $F_v$ renders the semantic and saliency evidence for the current observation. The record preserves defect semantics, localization, spatial support, and visual evidence in a common schema. It forms the interface to product-level reasoning.

\subsection{Multi-View Product Decision}

Because each observation is represented by the same evidence record, the per-view expert can be reused across the available views. For a product observed from $\mathcal V_N=\{1,\ldots,N\}$,
\begin{equation}
\mathcal E_v=f_{\rm expert}(I_v,P_{\rm det}),
\qquad v\in\mathcal V_N.
\label{eq:heads}
\end{equation}
Each view remains in its native image coordinate system because the observations are used to vary defect visibility rather than reconstruct a common geometry. Cross-view registration is therefore not required. The product-level VLM jointly considers the resulting records,
\begin{equation}
(\widehat y,\widehat c,\widehat{\mathcal V}_s)=
f_{\rm dec}\!\left(
\{\mathcal E_v\}_{v\in\mathcal V_N},
P_{\rm dec}
\right),
\label{eq:decision}
\end{equation}
where $\widehat y$ is the inspection verdict, $\widehat c$ is the predicted defect class, and $\widehat{\mathcal V}_s$ denotes the supporting observations. No majority rule is imposed. A defect supported by one observation can therefore contribute to the final verdict even when it is weak or absent in other views. This formulation separates view processing from product reasoning. Adding an observation requires another pass through the same expert and an additional evidence record, without changing the per-view model. The marginal benefit of additional optical observations is evaluated in Section~\ref{sec:ablation}.

After the visual verdict, the predicted class and inspection zone are used to retrieve the corresponding manufacturer acceptance rule and disposition. The retrieved information is attached to the supporting evidence for reporting and does not alter the visual decision.

\begin{figure}[t]
    \centering
    \smartfig[0.75\linewidth]{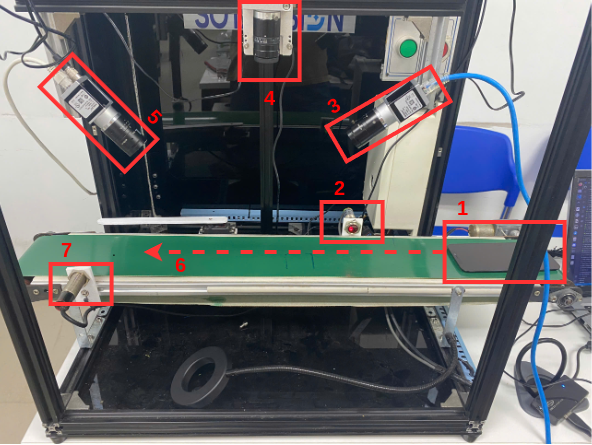}
    \caption{Production-line prototype used for industrial evaluation:
    (1) test product, (2,7) photoelectric sensors, (3--5) RGB cameras, and
    (6) conveyor direction. The three cameras provide nominal
    $45^\circ$, $90^\circ$, and $135^\circ$ observations.}
    \label{fig:realsetup}
\end{figure}

\section{Experiments}

\subsection{Common Protocol}

\textbf{Baselines.}
Direct semantic localization is evaluated with GPT-4o~\cite{openai2024gpt4o}, Gemini 3.6 Flash~\cite{gemini36flash2026}, Qwen2.5-VL-7B~\cite{hui2024qwen2}, and LLaVA-1.5-7B~\cite{liu2024improved}, with VT-ADL~\cite{mishra2021vtadl} included as an anomaly-localization reference. The complete per-view expert is compared with Faster R-CNN~\cite{ren2015faster}, RetinaNet~\cite{lin2017focal}, YOLOv8-L~\cite{varghese2024yolov8}, YOLOv11-L~\cite{khanam2024yolov11}, and YOLOv12-L~\cite{tian2025yolov12}. The supervised detectors use defect-box annotations, whereas the semantic branch is prompt-driven and the reconstruction branch is adapted using normal images. Supervised baselines are trained separately on the corresponding training split of each evaluation domain, with all reported test data kept disjoint. Their results therefore provide a supervised reference rather than a like-for-like training comparison.

\textbf{Metrics and implementation.}
Per-image localization is evaluated using class-wise AP$_{50}$ and standard
COCO AP. Direct VLMs do not expose detector-style confidence scores, so their
proposals are assigned a common fixed score and are used only as localization
references. Proposals from our expert are ranked by $o^*_{vj}$. Gemini 3.6 Flash
is used after the screening experiment below. The thresholds are fixed on a
disjoint validation set and kept unchanged for all reported tests at $\tau_a=0.60$,
$\tau_{\rm area}=2\times10^{-4}HW$, and $\tau_o=0.20$. Unless stated otherwise,
all localization metrics are computed before multi-view aggregation.

\subsection{Inspection Evaluation}

\textbf{SSGD protocol.}
SSGD is a public smartphone screen-glass defect dataset with seven annotated
categories \cite{han2023ssgd}. We sample a fixed five-class subset of
1,310 images from the original Crack, Broken, Blot, Spot, and Scratch
categories; Light-Leakage and Broken-Membrane are not included. A fixed
200-image subset of these 1,310 images is used only for VLM screening and
model selection; its results are therefore reported as a diagnostic rather
than an independent benchmark. As SSGD provides individual images rather than
repeated views of the same product, it evaluates only the per-view component.

\begin{figure*}[t]
\centering

\begin{subfigure}{0.48\textwidth}
    \centering
    \includegraphics[width=\linewidth]{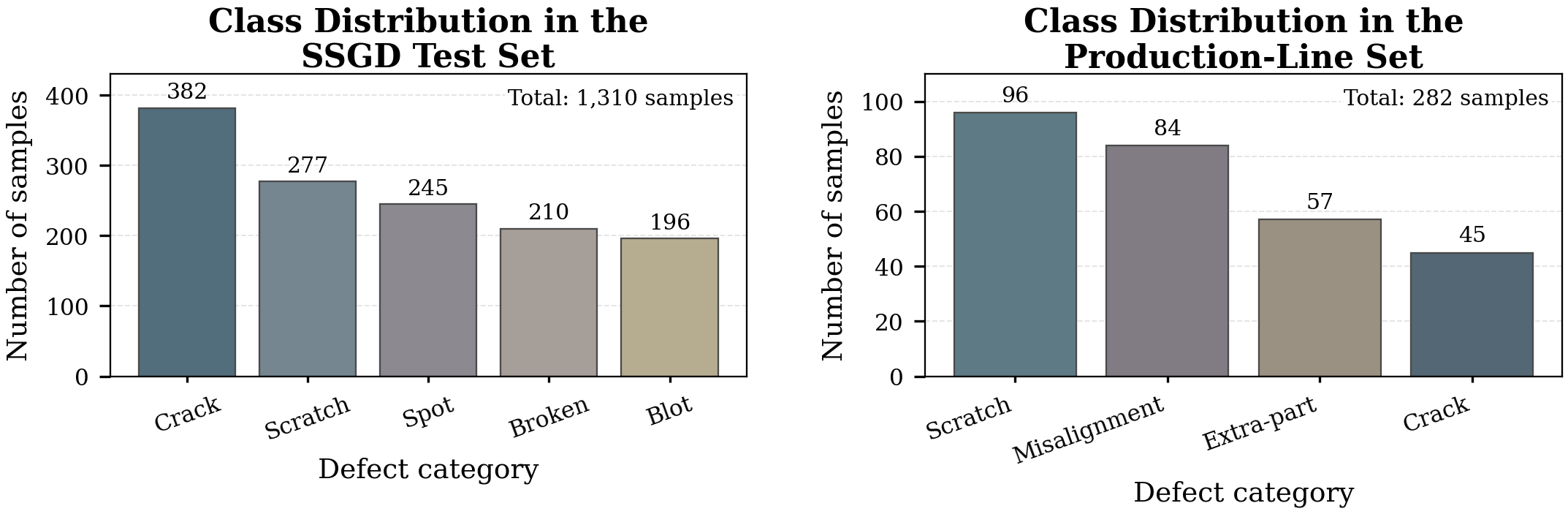}
    \caption{Class distributions of the five-class SSGD subset and the
    production-line set.}
\end{subfigure}
\hfill
\begin{subfigure}{0.49\textwidth}
    \centering
    \includegraphics[width=\linewidth]{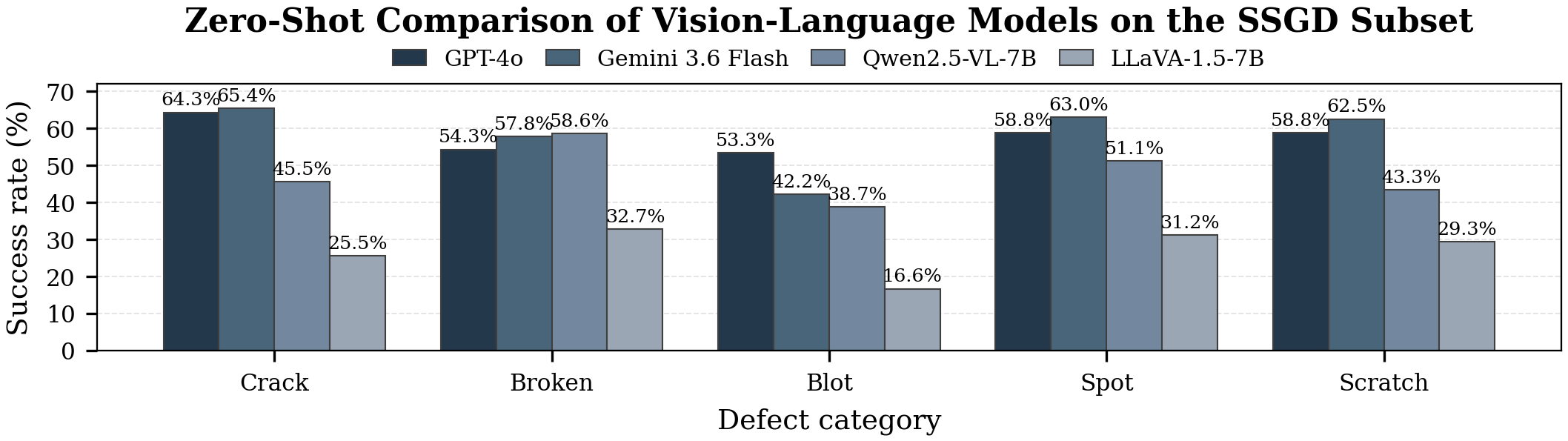}
    \caption{VLM screening on the 200-image SSGD subset.}
\end{subfigure}

\caption{Evaluation data and semantic-model screening. In (b), success denotes
a correct-class prediction with IoU $\geq0.5$; this statistic is used only for
model selection and is distinct from AP$_{50}$ in Table~\ref{tab:vlm200}.}
\label{fig:data}
\end{figure*}

\begin{table}[t]
\centering
\caption{AP$_{50}$ (\%) on the fixed 200-image SSGD screening subset. Mean is
the unweighted macro average.}
\label{tab:vlm200}
\footnotesize
\setlength{\tabcolsep}{0pt}
\renewcommand{\arraystretch}{1.10}
\begin{tabular*}{\columnwidth}{@{\extracolsep{\fill}}l cccccc@{}}
\toprule
\textbf{Method} & \textbf{Cra.} & \textbf{Bro.} & \textbf{Blot} &
\textbf{Spot} & \textbf{Scr.} & \textbf{Mean} \\
\midrule
VT-ADL  & 44.4 & 53.2 & 47.5 & 42.2 & 36.2 & 44.7 \\
LLaVA-1.5-7B                  & 25.5 & 32.7 & 16.6 & 31.2 & 29.3 & 27.1 \\
Qwen2.5-VL-7B                 & 43.5 & 53.1 & 38.2 & 51.5 & 43.1 & 45.9 \\
GPT-4o                        & 61.2 & 50.3 & 50.0 & 51.8 & 49.9 & 52.6 \\
Gemini 3.6 Flash              & 63.4 & 54.7 & 43.2 & \textbf{64.0} & 60.0 & 57.1 \\
\midrule
\textbf{\Ours{} 1-view} & \textbf{74.1} & \textbf{62.0} & \textbf{58.8} &
60.5 & \textbf{67.5} & \textbf{64.6} \\
\bottomrule
\end{tabular*}
\end{table}

Table~\ref{tab:vlm200} first identifies the semantic model used in the remaining
experiments. Gemini 3.6 Flash gives the strongest direct-VLM mean AP$_{50}$ at
57.1\%, while saliency-based ranking raises AP$_{50}$ to 64.6\%. The gain is not uniform: Spot decreases from 64.0\% to 60.5\%,
supporting the use of saliency as soft ranking evidence rather than a hard
acceptance rule. The full 1,310-image protocol then compares the per-view expert
with supervised detectors when no multi-view information is available. In
Table~\ref{tab:detpub}, YOLOv12-L gives the highest AP and AP$_{50}$, while our
expert gives the highest AP$_{75}$ and AP$_S$. The overall AP difference is
small, 40.4\% versus 39.8\%, so the result supports competitive single-view
localization rather than superiority over supervised detection.

\begin{table}[t]
\centering
\caption{COCO detection metrics (\%) on the 1,310-image SSGD protocol. All rows
are single-view.}
\label{tab:detpub}
\footnotesize
\setlength{\tabcolsep}{0pt}
\renewcommand{\arraystretch}{1.10}
\begin{tabular*}{\columnwidth}{@{\extracolsep{\fill}}l cccccc@{}}
\toprule
\textbf{Method} & \textbf{AP} & \textbf{AP$_{50}$} & \textbf{AP$_{75}$} &
\textbf{AP$_S$} & \textbf{AP$_M$} & \textbf{AP$_L$} \\
\midrule
Faster R-CNN   & 22.6 & 48.3 & 24.5 & 12.4 & 23.8 & 30.5 \\
RetinaNet      & 28.4 & 56.7 & 30.6 & 16.8 & 29.7 & 36.9 \\
YOLOv8-L       & 38.2 & 68.4 & 40.7 & 27.4 & 39.4 & 48.2 \\
YOLOv11-L      & 39.1 & 69.5 & 41.7 & 28.2 & 40.4 & 49.3 \\
YOLOv12-L      & \textbf{40.4} & \textbf{70.9} & 43.2 & 29.4 &
\textbf{41.9} & \textbf{50.7} \\
\Ours{} 1-view & 39.8 & 68.7 & \textbf{44.0} & \textbf{30.3} & 41.2 & 47.8 \\
\bottomrule
\end{tabular*}
\end{table}

\textbf{Production-line protocol.}
The production-line prototype in Fig.~\ref{fig:realsetup} contains three fixed
RGB cameras configured at nominal left-oblique, frontal, and right-oblique
orientations of $45^\circ$, $90^\circ$, and $135^\circ$, respectively. The
industrial set contains 94 defective products: 32 Scratch, 28 Misalignment, 19 Extra-part, and 15 Crack, yielding 282 images at $1944\times2592$.

During acquisition, each product is held stationary while the three views are captured sequentially. After inference, the conveyor resumes and routes the product according to the inspection verdict. The system uses an Intel Core i9-14900 CPU with 32 threads and 16~GB RAM. The current rig provides three viewpoints, so we evaluate all configurations with $N\in\{1,2,3\}$. Larger $N$ is not considered because it requires additional hardware, expert passes, and product-level context. The 282 images are first evaluated independently to isolate per-view performance before analyzing view complementarity.

Table~\ref{tab:vlmreal} tests whether semantic--saliency association remains
effective under the production imaging conditions. The strongest direct VLM
reaches 52.6\% macro AP$_{50}$, while the complete expert reaches 62.6\% and
improves all four defect classes. The ablation below isolates whether this gain
comes from evidence association rather than a change in semantic localization.

\begin{table}[t]
\centering
\caption{Class-wise AP$_{50}$ (\%) on 282 production-line images. Results are
per image and exclude product-level aggregation.}
\label{tab:vlmreal}
\footnotesize
\setlength{\tabcolsep}{0pt}
\renewcommand{\arraystretch}{1.10}
\begin{tabular*}{\columnwidth}{@{\extracolsep{\fill}}l ccccc@{}}
\toprule
\textbf{Method} & \textbf{Cra.} & \textbf{Ext.} & \textbf{Mis.} &
\textbf{Scr.} & \textbf{Mean} \\
\midrule
VT-ADL  & 32.4 & 52.4 & 44.5 & 40.8 & 42.5 \\
LLaVA-1.5-7B                  & 21.1 & 24.6 &  9.6 & 24.4 & 19.9 \\
Qwen2.5-VL-7B                 & 47.6 & 50.9 & 41.5 & 43.6 & 45.9 \\
GPT-4o                        & 49.1 & 51.0 & 48.6 & 51.0 & 49.9 \\
Gemini 3.6 Flash              & 60.1 & 52.8 & 41.8 & 55.6 & 52.6 \\
\midrule
\textbf{\Ours}              & \textbf{70.5} & \textbf{61.8} &
\textbf{55.0} & \textbf{62.9} & \textbf{62.6} \\
\bottomrule
\end{tabular*}
\end{table}

Table~\ref{tab:detreal} places the same per-view expert against supervised
detectors on the production-line images. Our method is higher in AP,
AP$_{75}$, AP$_S$, and AP$_M$, while YOLOv12-L remains higher at AP$_{50}$ and
AP$_L$. The margins are small and do not establish overall superiority. This establishes competitive per-view localization under the target imaging conditions before the multi-view study.

\begin{table}[t]
\centering
\caption{COCO detection metrics (\%) on 282 production-line images.}
\label{tab:detreal}
\footnotesize
\setlength{\tabcolsep}{0pt}
\renewcommand{\arraystretch}{1.10}
\begin{tabular*}{\columnwidth}{@{\extracolsep{\fill}}l cccccc@{}}
\toprule
\textbf{Method} & \textbf{AP} & \textbf{AP$_{50}$} & \textbf{AP$_{75}$} &
\textbf{AP$_S$} & \textbf{AP$_M$} & \textbf{AP$_L$} \\
\midrule
Faster R-CNN & 18.9 & 43.4 & 20.1 & 10.1 & 19.8 & 25.7 \\
RetinaNet    & 24.1 & 50.0 & 25.8 & 13.6 & 25.1 & 31.5 \\
YOLOv8-L     & 33.6 & 61.8 & 35.8 & 23.6 & 35.1 & 42.9 \\
YOLOv11-L    & 34.9 & 63.1 & 37.1 & 24.8 & 36.4 & 44.2 \\
YOLOv12-L    & 36.1 & \textbf{64.7} & 38.8 & 26.3 & 37.8 & \textbf{45.7} \\
\Ours        & \textbf{37.2} & 62.6 & \textbf{39.8} & \textbf{27.6} &
\textbf{38.4} & 44.9 \\
\bottomrule
\end{tabular*}
\end{table}

\subsection{Ablation and Multi-View Evaluation}
\label{sec:ablation}

\textbf{Evidence association.}
Table~\ref{tab:evidence} isolates the proposed within-view association by
holding every semantic box fixed. Semantic proposals alone obtain 52.6\%
AP$_{50}$. Ranking the same proposals by their saliency response increases the
score to 58.6\%, and explicit semantic--saliency region association reaches
62.6\%. Since the predicted coordinates are identical across all three
settings, the 10.0-point gain is attributable to evidence-based ranking rather
than geometric refinement. The additional 4.0 points from region association
further suggest that explicit spatial agreement is more informative here than
the mean saliency response inside a VLM-selected box.

\begin{table}[t]
\centering
\caption{Evidence ablation on 282 production-line images. Semantic coordinates
are fixed in all rows.}
\label{tab:evidence}
\small
\setlength{\tabcolsep}{2.4pt}
\renewcommand{\arraystretch}{1.12}
\begin{tabular*}{\columnwidth}{@{\extracolsep{\fill}}l cccc@{}}
\toprule
\textbf{Representation} & \textbf{Sem.} & \textbf{Heat} & \textbf{Boxes} &
\textbf{AP$_{50}$} \\
\midrule
Semantic only       & \cmark & \xmark & \xmark & 52.6 \\
+ saliency response & \cmark & \cmark & \xmark & 58.6 \\
+ saliency boxes    & \cmark & \cmark & \cmark & \textbf{62.6} \\
\bottomrule
\end{tabular*}
\end{table}

\textbf{Optical view complementarity.}
We evaluate all one-, two-, and three-view subsets to determine whether the
available viewpoints expose complementary defect evidence. A product contributes
to $R_{\rm prod}@0.5$ when at least one selected view contains a correct-class
prediction with IoU $\geq0.50$. Because cross-verification neither changes
nor discards semantic proposals, this metric isolates optical view
complementarity and does not evaluate the product-level VLM.

Table~\ref{tab:views} shows that no single viewpoint dominates all defect
classes. The best single view reaches 75.5\%, the best pair 85.1\%, and all
three views 88.3\%. The second view contributes 9.6 percentage points over the
best single view, while the third contributes a further 3.2 points. The smaller
increment indicates diminishing returns within the tested acquisition range,
highlighting the trade-off between additional optical coverage and the added
acquisition and processing load.

\begin{table}[t]
\centering
\caption{$R_{\rm prod}@0.5$ (\%) for all available view subsets.}
\label{tab:views}
\footnotesize
\setlength{\tabcolsep}{0pt}
\renewcommand{\arraystretch}{1.08}
\begin{tabular*}{\columnwidth}{@{\extracolsep{\fill}}l ccccc@{}}
\toprule
\textbf{Views} & \textbf{Cra.} & \textbf{Ext.} & \textbf{Mis.} &
\textbf{Scr.} & \textbf{All} \\
\midrule
Frontal & 66.7 & 73.7 & \textbf{78.6} & 68.8 & 72.3 \\
Left    & 73.3 & \textbf{78.9} & 75.0 & \textbf{75.0} & 75.5 \\
Right   & 73.3 & 73.7 & \textbf{78.6} & 71.9 & 74.5 \\
\midrule
F + L   & 80.0 & 84.2 & 82.1 & 81.2 & 81.9 \\
F + R   & 80.0 & 84.2 & 82.1 & 78.1 & 80.9 \\
L + R   & 86.7 & 84.2 & 85.7 & 84.4 & 85.1 \\
\midrule
\textbf{F + L + R} & \textbf{86.7} & \textbf{89.5} & \textbf{89.3} &
\textbf{87.5} & \textbf{88.3} \\
\bottomrule
\end{tabular*}
\end{table}

\begin{figure}[t] 
    \centering

    \begin{subfigure}[t]{0.32\columnwidth}
        \centering
        \includegraphics[height=1.60cm]{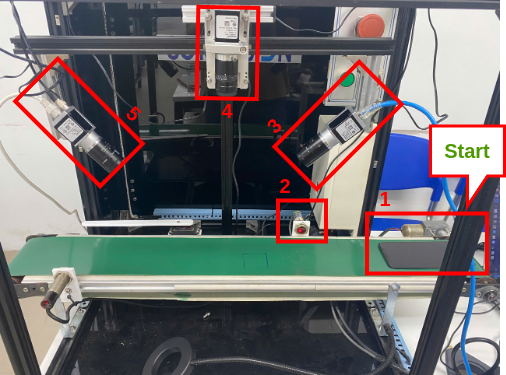}
        \caption{Frontal.}
        \label{fig:crack_rgb_front}
    \end{subfigure}\hfill
    \begin{subfigure}[t]{0.32\columnwidth}
        \centering
        \includegraphics[height=1.60cm]{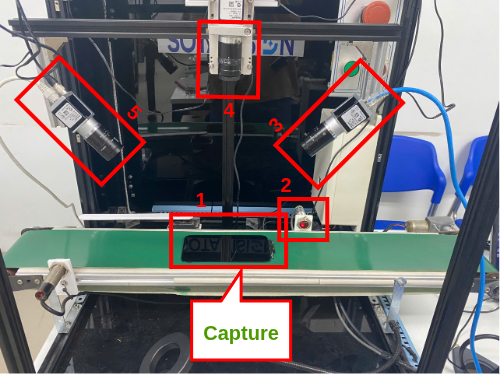}
        \caption{Left-oblique.}
        \label{fig:crack_rgb_left}
    \end{subfigure}\hfill
    \begin{subfigure}[t]{0.32\columnwidth}
        \centering
        \includegraphics[height=1.60cm]{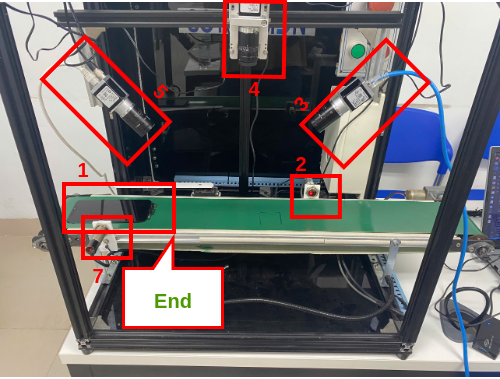}
        \caption{Right-oblique.}
        \label{fig:crack_rgb_right}
    \end{subfigure}

    \vspace{1.0mm}

    \begin{subfigure}[t]{0.32\columnwidth}
        \centering
        \includegraphics[height=1.20cm]{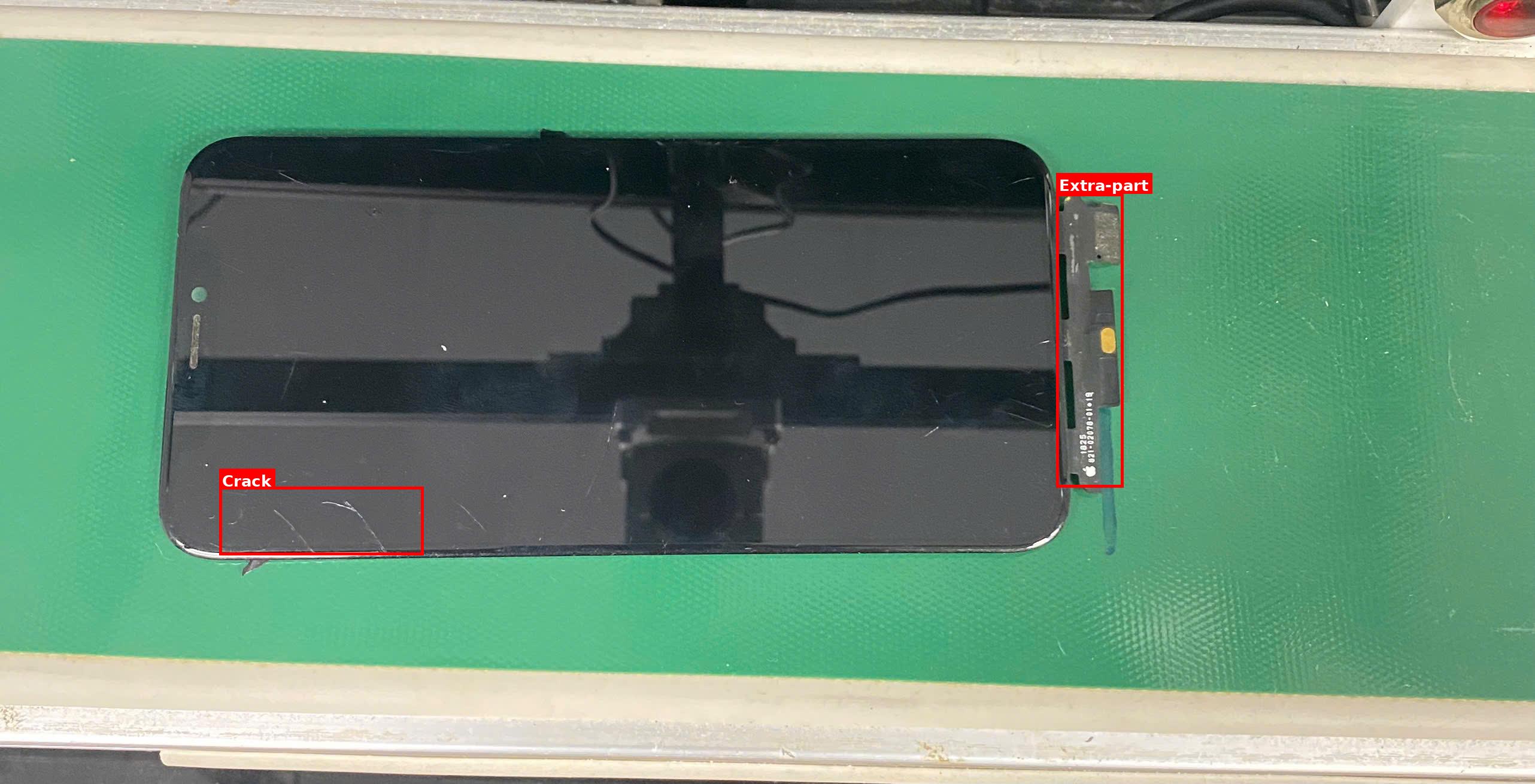}
        \caption{Det., front.}
        \label{fig:crack_det_front}
    \end{subfigure}\hfill
    \begin{subfigure}[t]{0.32\columnwidth}
        \centering
        \rotatebox{90}{
            \includegraphics[width=1.20cm]{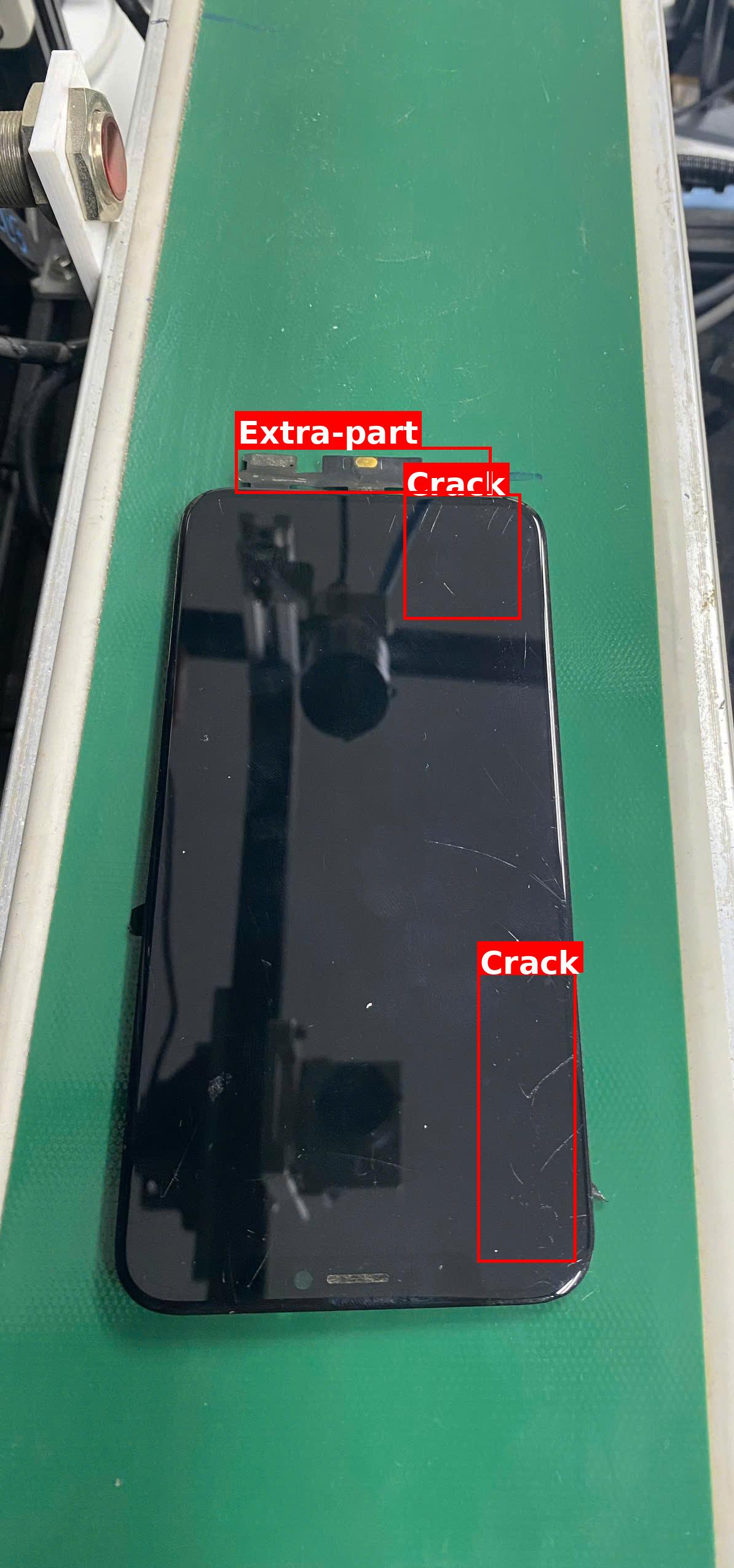}
        }
        \caption{Det., left.}
        \label{fig:crack_det_left}
    \end{subfigure}\hfill
    \begin{subfigure}[t]{0.32\columnwidth}
        \centering
        \rotatebox{90}{
            \includegraphics[width=1.20cm]{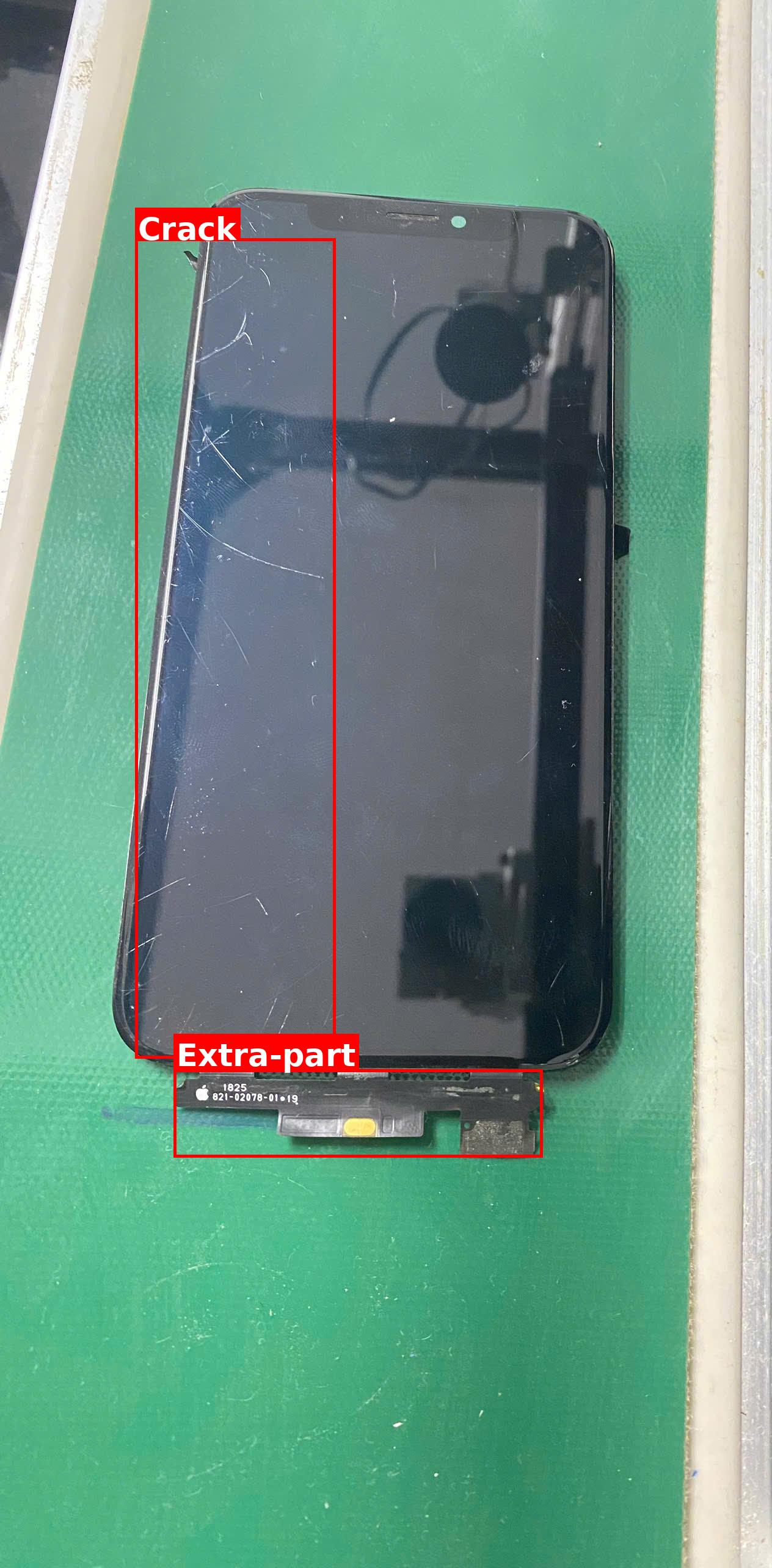}
        }
        \caption{Det., right.}
        \label{fig:crack_det_right}
    \end{subfigure}

    \vspace{1.0mm}

    \begin{subfigure}[t]{0.32\columnwidth}
        \centering
        \includegraphics[height=1.20cm]{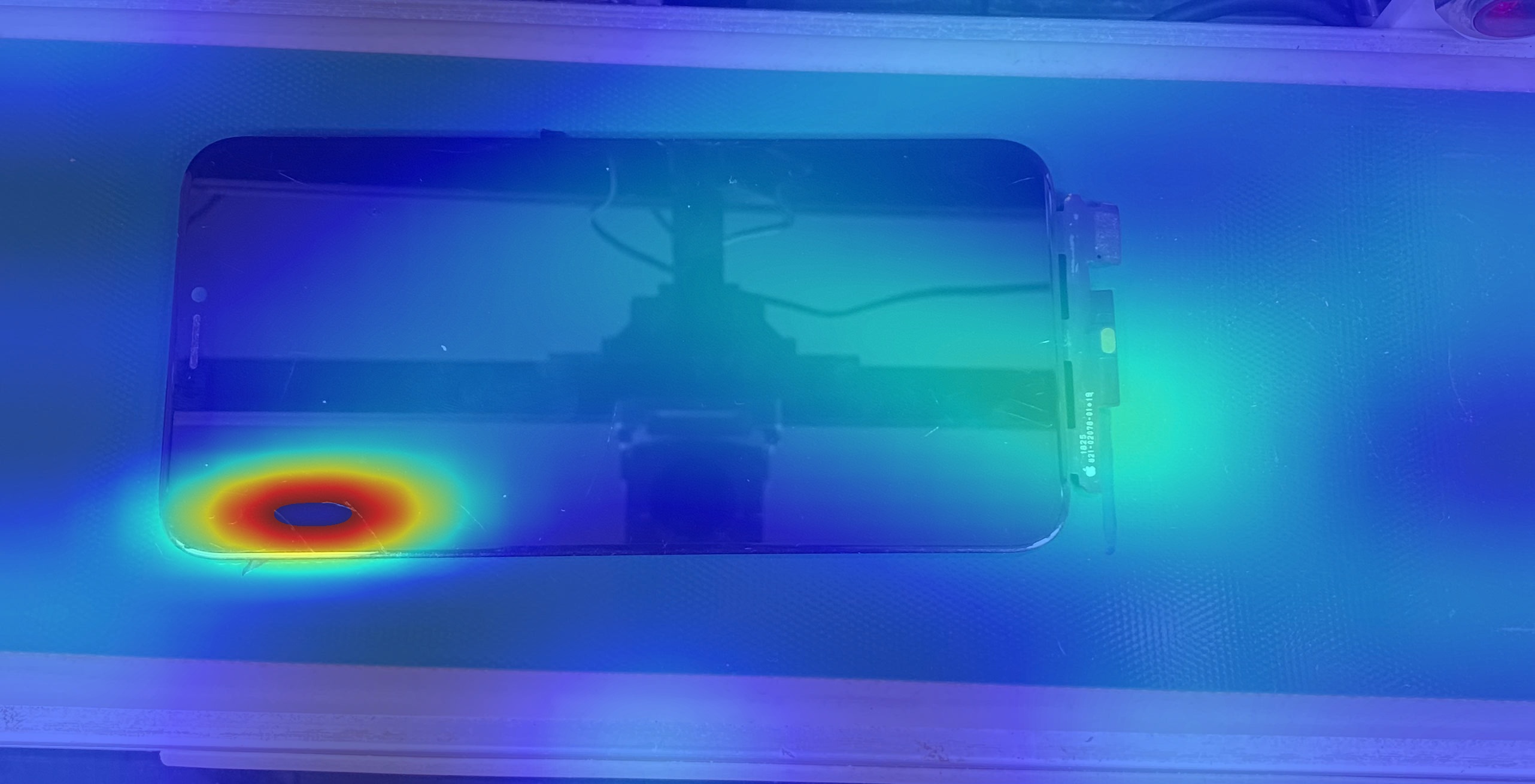}
        \caption{Sal., front.}
        \label{fig:crack_sal_front}
    \end{subfigure}\hfill
    \begin{subfigure}[t]{0.32\columnwidth}
        \centering
        \rotatebox{90}{
            \includegraphics[width=1.20cm]{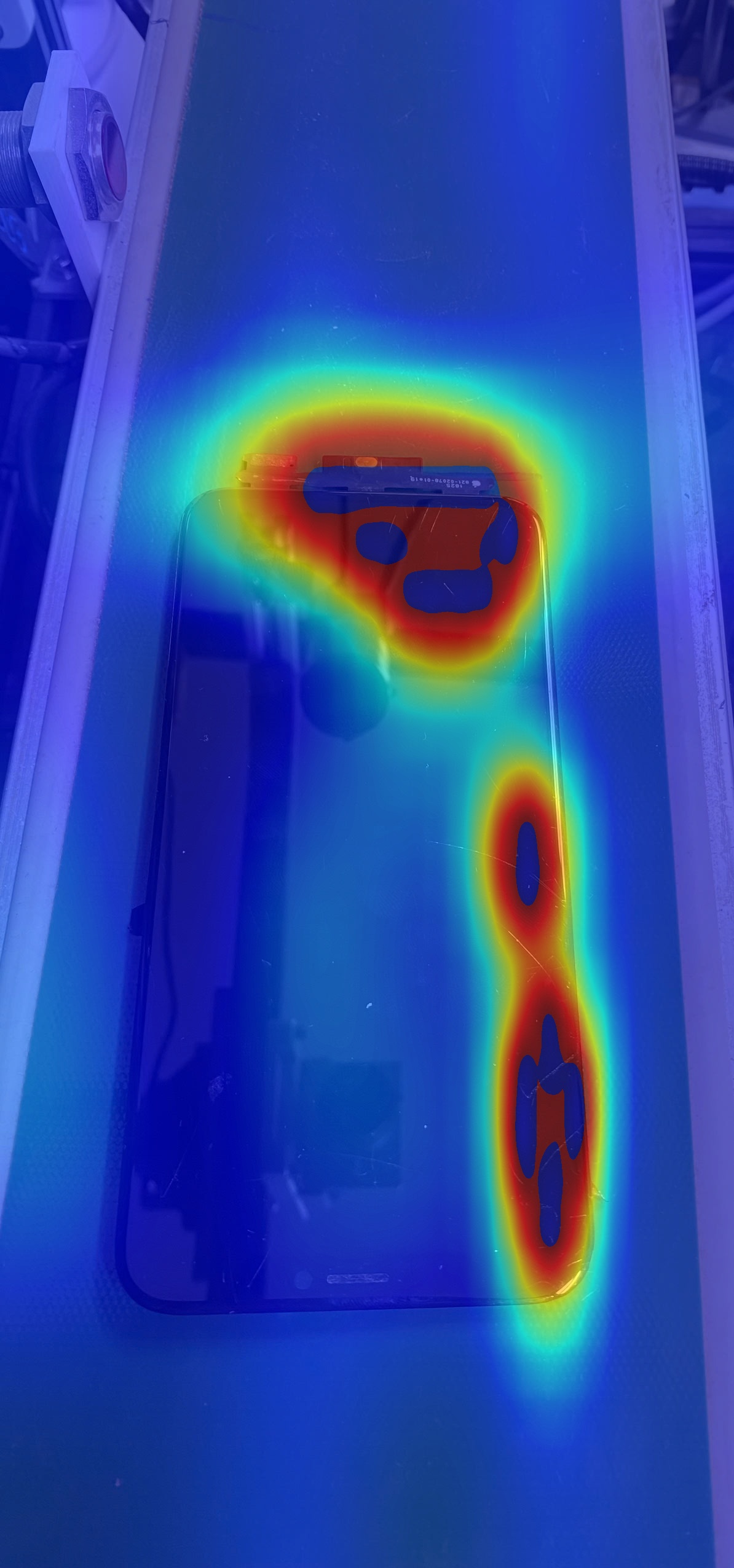}
        }
        \caption{Sal., left.}
        \label{fig:crack_sal_left}
    \end{subfigure}\hfill
    \begin{subfigure}[t]{0.32\columnwidth}
        \centering
        \rotatebox{90}{
            \includegraphics[width=1.20cm]{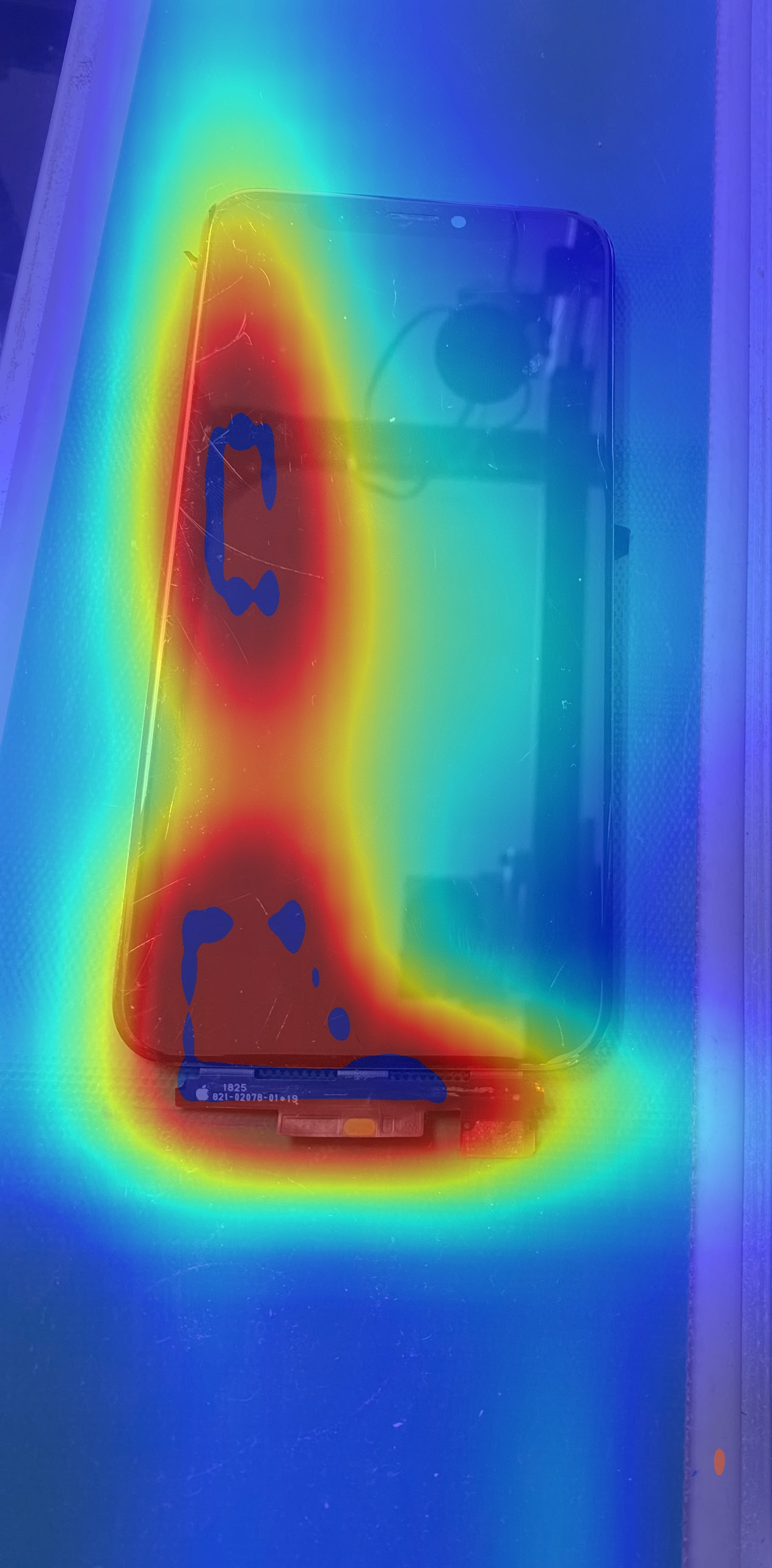}
        }
        \caption{Sal., right.}
        \label{fig:crack_sal_right}
    \end{subfigure}

    \vspace{1.0mm}

    \begin{subfigure}[t]{0.32\columnwidth}
        \centering
        \includegraphics[height=1.20cm]{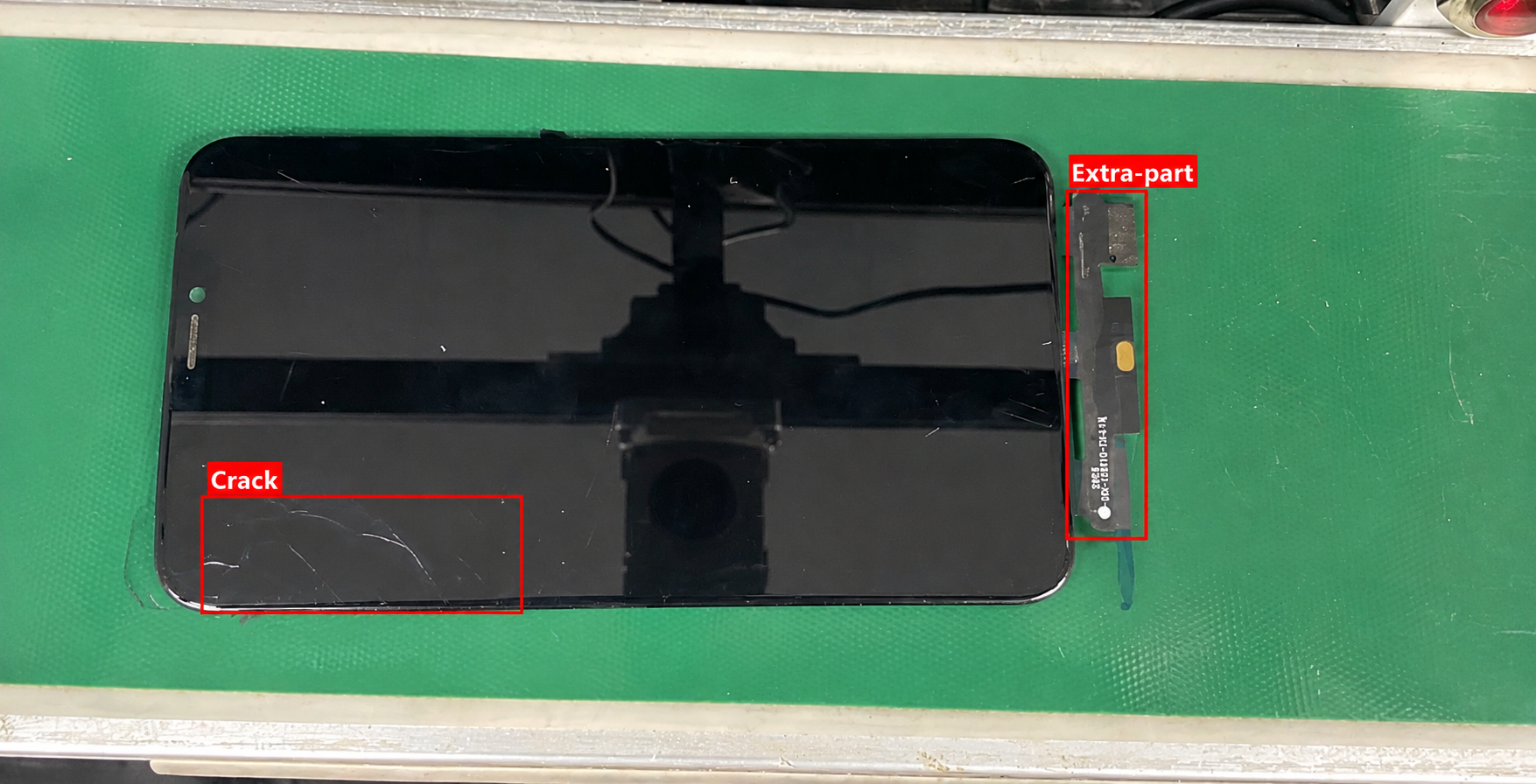}
        \caption{Evidence, front.}
        \label{fig:crack_evidence_front}
    \end{subfigure}\hfill
    \begin{subfigure}[t]{0.32\columnwidth}
        \centering
        \rotatebox{90}{%
            \IfFileExists{figures/extra_left_fused.png}{%
                \includegraphics[width=1.20cm]{figures/extra_left_fused.png}%
            }{%
                \includegraphics[width=1.20cm]{figures/extra_left_detect.jpg}%
            }%
        }
        \caption{Evidence, left.}
        \label{fig:crack_evidence_left}
    \end{subfigure}\hfill
    \begin{subfigure}[t]{0.32\columnwidth}
        \centering
        \rotatebox{90}{
            \includegraphics[width=1.20cm]{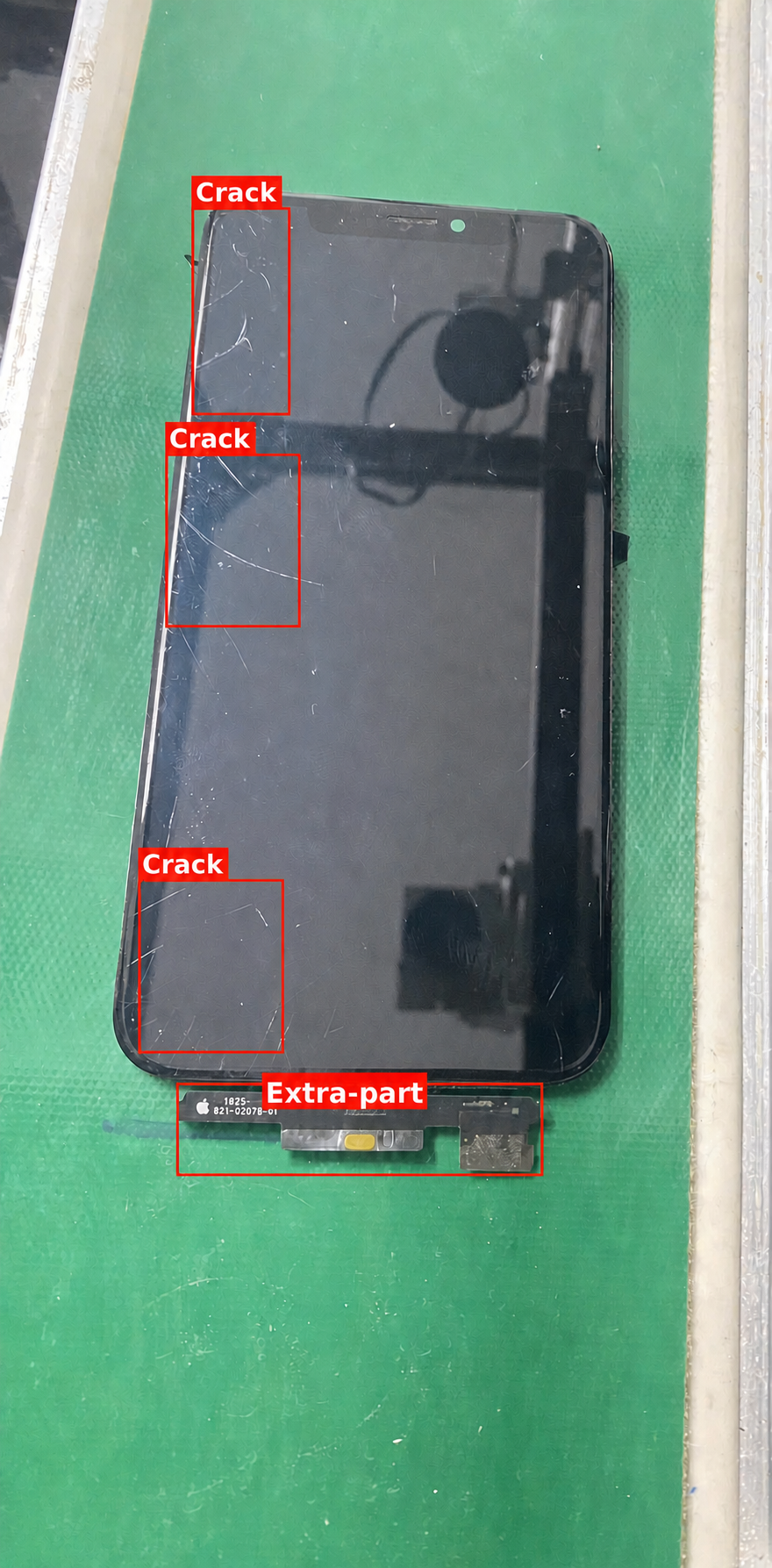}
        }
        \caption{Evidence, right.}
        \label{fig:crack_evidence_right}
    \end{subfigure}

    \vspace{1.0mm}

    \caption{
        Representative extra-part product across the three viewpoints.
        Semantic localization and normal-reference saliency are shown with the
        resulting cross-verified evidence. Spatial association changes proposal
        support while preserving the semantic coordinates.
    }
    \label{fig:qual}
\end{figure}

Fig.~\ref{fig:qual} complements the quantitative ablation. Semantic localization
is preserved while saliency changes the support attached to each proposal.
Spatial agreement is informative but not a correctness certificate, since
reflection can affect both cues.

\begin{figure}[t]
    \centering

    \begin{subfigure}[t]{0.315\columnwidth}
        \centering
        \includegraphics[height=1.65cm,keepaspectratio]{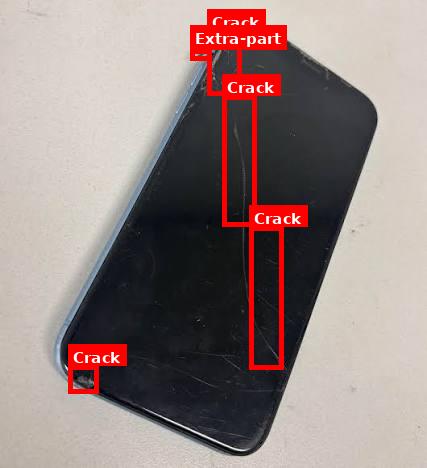}
        \caption{Example 1.}
        \label{fig:ood1}
    \end{subfigure}
    \hfill
    \begin{subfigure}[t]{0.315\columnwidth}
        \centering
        \includegraphics[height=1.65cm,keepaspectratio]{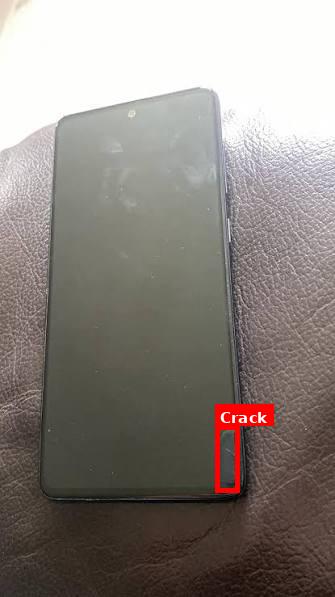}
        \caption{Example 2.}
        \label{fig:ood2}
    \end{subfigure}
    \hfill
    \begin{subfigure}[t]{0.315\columnwidth}
        \centering
        \includegraphics[height=1.65cm,keepaspectratio]{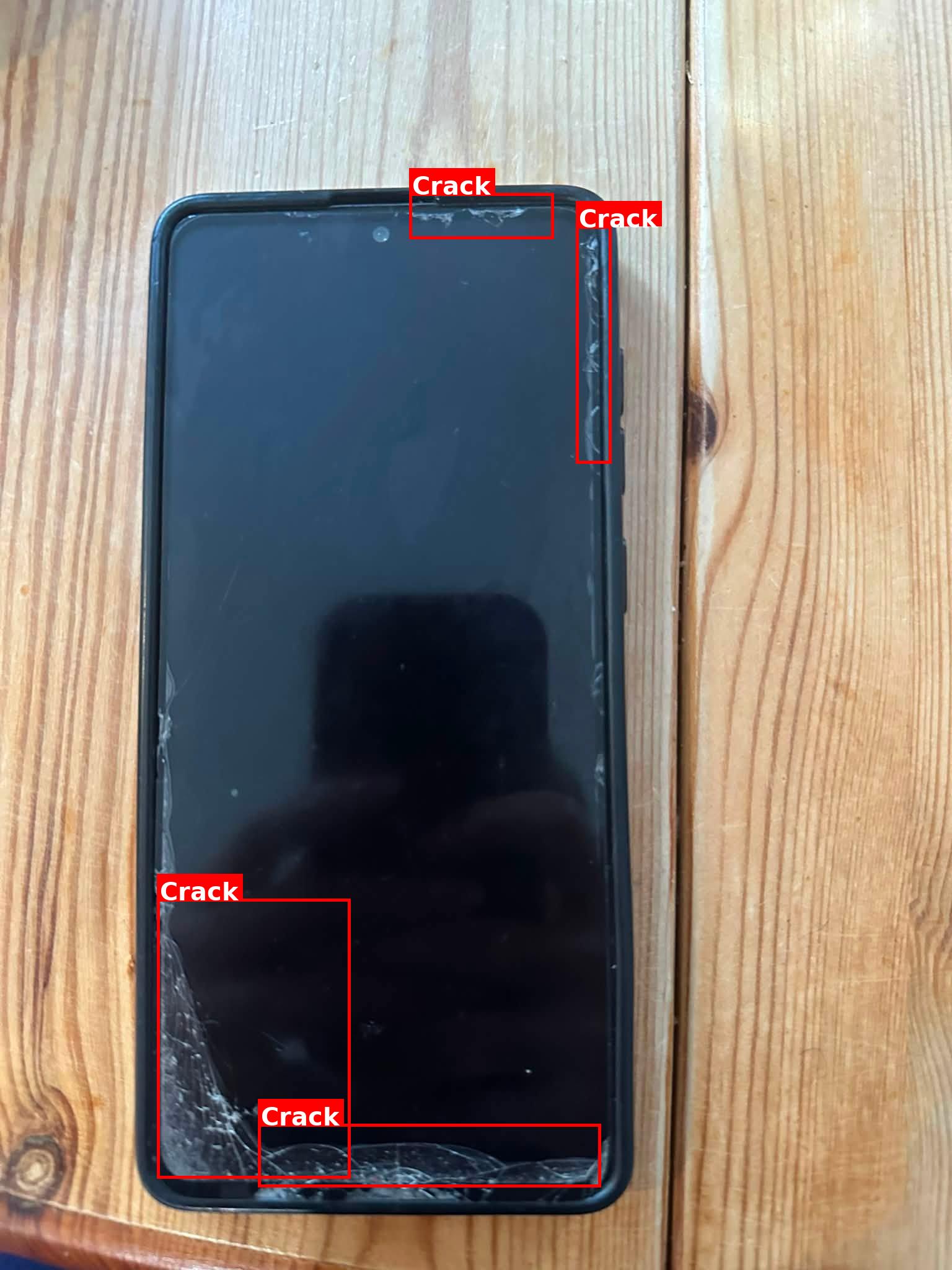}
        \caption{Example 3.}
        \label{fig:ood3}
    \end{subfigure}

    \caption{Cross-verified predictions on three images outside the evaluation
    datasets, exhibiting different object appearances and imaging conditions.}
    \label{fig:ood}
\end{figure}

Fig.~\ref{fig:ood} further illustrates the cross-verification behavior under
appearance conditions outside the evaluation datasets. The observed spatial
agreement between semantic proposals and normal-reference responses is
consistent with the evidence-association result in Table~\ref{tab:evidence},
although these examples are qualitative and do not establish OOD
generalization.
\section{Conclusion}

We presented a multi-view inspection framework for reflective smartphone cover glass. A shared per-view expert associates VLM semantic localization with
normal-reference saliency through spatial support, while the resulting evidence
records are combined at product level without cross-view registration. On
production-line data, semantic--saliency association improves AP$_{50}$ from
52.6\% to 62.6\% with fixed semantic coordinates, while complementary optical
views increase $R_{\rm prod}@0.5$ from 75.5\% for the best single view to
88.3\% using all three views. These results support the complementary roles of
within-view evidence association and multi-view optical diversity.

The current evaluation is limited to three physical viewpoints, one product
family, and a production set containing defective products only. The
product-level VLM is also not evaluated independently from evidence
availability across views. Future work will extend the evaluation to normal
products, broader production conditions, and additional acquisition
configurations, with direct assessment of product-level decision reliability.

\bibliographystyle{IEEEtran}
\bibliography{refs}

\end{document}